\RequirePackage[T1]{fontenc}
\documentclass[letterpaper,10pt,conference]{ieeeconf}
\IEEEoverridecommandlockouts
\usepackage{amsmath,amssymb,array,graphicx,booktabs,flushend,cite,microtype,times}
\usepackage[hidelinks]{hyperref}
\usepackage{float}
\newcommand{\bank}{\mathcal{B}}
\newcommand{\mem}{\mathcal{M}}
\newcommand{\pos}{\mathcal{P}}
\newcommand{\negset}{\mathcal{N}}
\newcommand{\ind}{\mathbf{1}}
\title{\LARGE\bf Sequential Recovery Updates for Learned Controllers\\under Successful-Trajectory Preservation}
\author{Zongtan Li$^{1}$, Tianming Liu$^{1}$, Ye Zhao$^{2}$\\
$^{1}$University of Georgia, Athens, GA, USA\\
$^{2}$Georgia Institute of Technology, Atlanta, GA, USA}
\begin{document}
\maketitle
\thispagestyle{empty}\pagestyle{empty}

\begin{abstract}
A competent learned controller already recovers from many disturbances. Improving it means learning a correction, and learning when not to apply it. Freezing a skill's parameters does not settle the second half. In a controller that routes among frozen skills, a new entry rule can redirect the states an old skill succeeded on. We therefore constrain entry instead of parameters. Each update searches for a feedback skill at a measured failure. It fits an entry rule that reads observations and rejects every state on an accumulated successful trajectory. The skill is admitted only if the composed controller reproduces the new recovery on its own and replays every protected trajectory exactly. A finite-trajectory condition states when this preserves the previous controller. One mechanism runs on four streams: two motions on each of two humanoid systems. Two updates per stream add 62 recoveries over 192 withheld perturbations, and lose none of the base controller's successes. We then price that guarantee. Removing the constraint recovers 33 more conditions but stops eight protected ones from completing at all, and excluding only what admission proves necessary recovers none of that cost.
\end{abstract}

\section{Introduction}
An existing learned feedback controller can already provide competent
closed-loop behavior, making adaptation a problem of selective improvement.
Humanoid motion tracking illustrates this setting through increasingly
broad learned repertoires
\cite{merel2019npmp,wagener2022mocapact,peng2018deepmimic,luo2026sonic}. A disturbance does not
necessarily require a new action: the original controller may already
recover. A learned correction that improves a difficult failure can disrupt
a nearby successful response. Repeatedly replacing a shared recovery policy
can also erase a skill acquired at an earlier disturbance. The useful
objective is consequently to \emph{accumulate recoveries while retaining
behavior that already works}.

Consider such a controller as one component of a stack validated as a whole.
An update is acceptable there only if one can state precisely what it
changed, and retraining cannot: gradient updates to shared parameters move
every trajectory a little, with no enumeration of what moved, so every
downstream test must be repeated. Replaying old successes into the loss
reduces that movement without removing it. We ask instead for a checkable
property: on a recorded set of successful episodes, the updated controller
must reproduce the identical trajectory. The difficult part is then composing
the skills. Freezing their parameters prevents parameter interference, but a
new entry rule can still select the wrong skill and change an old successful
trajectory, so behavioral retention is a property of the entire closed loop,
including skill selection.

\begin{figure*}[t]
\centering
\includegraphics[width=.98\textwidth]{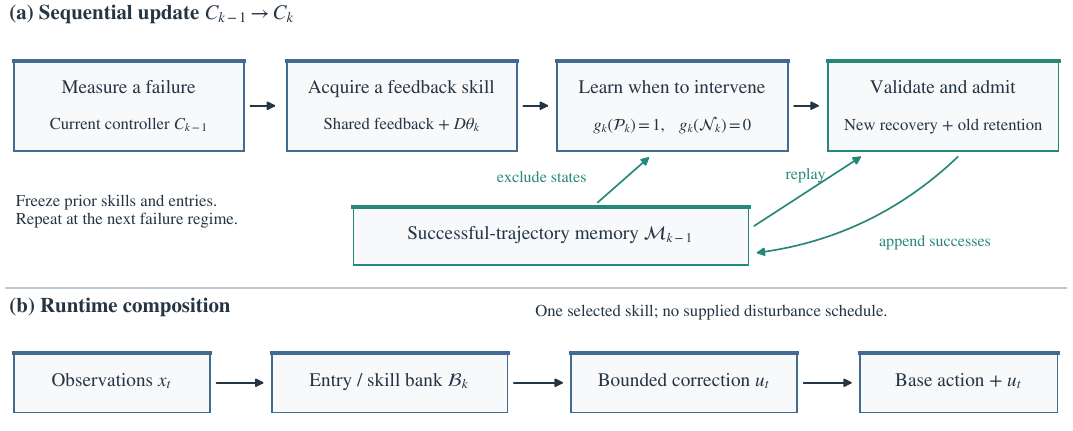}
\caption{Sequential acquisition and preservation: (a) updates between episodes,
(b) runtime composition from observations. Accumulated successful histories
constrain each new entry, and closed-loop recovery and retention tests
decide admission. Runtime selection receives no push schedule.}
\label{fig:method}
\end{figure*}

We address this problem by coupling acquisition driven by measured failures with
\emph{successful-trajectory exclusion}. At each stage we measure a
success/failure bracket of the current composed controller, search for a
feedback recovery under a fixed budget, and learn an entry rule that
recognizes demonstrated recovery onsets while rejecting states recorded on
successful histories. We append the skill only when autonomous recovery and
retention measured by replay pass. Accepted successful experience becomes a
constraint on the next update. At runtime, the base controller remains the
default outside admitted intervention windows, and observations select
which skill receives correction authority (Fig.~\ref{fig:method}).

The contribution is the connection between skill acquisition, autonomous
composition, and recorded-behavior preservation. Residual actions, search,
frozen expert banks, and linear classification are existing tools;
none alone solves the composition problem. We make three contributions:
\begin{enumerate}
\item A sequential recovery mechanism that couples measured failures,
an entry rule read from observations and constrained by successful
histories, and a closed-loop admission test that enforces exact reproduction of recorded
behavior.
\item A finite-trajectory preservation condition explaining why unchanged
expert parameters are insufficient and when excluding a new entry preserves
the old composed controller exactly.
\item A measurement of what that guarantee costs, over four streams and two
systems: 33 recoveries forgone, against 11 withheld and eight protected
conditions saved and 46 of 83 recorded trajectories held identical. A
control excluding only what admission proves necessary returns none of that
cost, so it is intrinsic to the guarantee rather than to our
implementation.
\end{enumerate}
The formulation does not depend on humanoid morphology; we evaluate it on
humanoid motion tracking as a demanding balance-and-contact setting. The
experiments establish short sequential streams in fixed simulation systems,
not indefinite online physical learning, universal absence of regression, or
transfer of a skill between plants.

\section{Related Work}
Classical push recovery studies capture and balance constraints
\cite{pratt2006capture,koolen2012capturability}; learned humanoid controllers
can express richer whole-body responses. Residual policy learning and
residual reinforcement learning augment an existing controller
\cite{silver2018residual,johannink2019residual}. Recovery RL learns recovery
zones for switching between task and recovery behavior
\cite{thananjeyan2021recovery}. These works motivate selective correction,
but a recovery region alone does not constrain a new intervention to
preserve the successful trajectories of the controller it augments.
Our contribution concerns the acquisition and admission of successive
recoveries into such a controller.

Cerebellar-inspired residual adaptation already combines a frozen policy,
local correction, and consolidation \cite{jayasinghe2026cerebellar}, so we
claim no priority for continual residual adaptation. Our focus is an
explicit successful-history constraint on autonomous entry, a stage-by-stage
test of acquisition against retention, and a measurement of what that
constraint costs.

Progressive networks preserve prior parameters by adding capacity
\cite{rusu2016progressive}, which addresses interference inside a learned
representation. Preserving parameters does not preserve a routed control
trajectory, however, because a new selector can alter which fixed component
acts; our entry constraint targets that second source of interference, at
the price of a growing skill and trajectory memory.

Simplex architectures separate high-performance behavior from an assurance
mechanism, and Neural Simplex also supports adaptation
\cite{sha2001simplex,phan2020neuralsimplex}; safe policy improvement with
baseline bootstrapping gives guarantees under different offline assumptions
\cite{laroche2019spibb}. We do not equate a learned entry classifier with a
certified safety filter: our preservation statement covers stored
deterministic trajectories, and gains or losses elsewhere are measurements.

Stress testing searches for system-specific failures
\cite{corso2021survey}, while curricula adapt training difficulty
\cite{florensa2017reverse,dennis2020paired}. We query the controller for
failures to choose where a recovery is needed, obtaining a measured bracket
rather than a proof of a monotone boundary. Such magnitudes tell an update
where to train but nothing about what it must leave intact, and only the
second question decides whether successive updates accumulate.
The outcome we evaluate is therefore cumulative recovery at held-out
perturbations and retention of earlier competence, rather than an enlarged
disturbance range.

\section{Sequential Recovery with Successful Histories}
\subsection{Problem Setting and the Update Contract}
Consider a plant with state $s_t$, action $a_t$, and transition
$s_{t+1}=F(s_t,a_t,d_t)$ under disturbance $d_t$. An existing learned
feedback controller $\pi_0$ supplies the nominal action. Its internal state
and weights need not be differentiable through the plant. A task condition
$q$ specifies a task, reset, and disturbance realization; $Y(C,q)\in\{0,1\}$
denotes successful completion by composed controller $C$. We assume an
observable runtime interface and a resettable evaluation environment.
Updates occur between episodes, with the plant and base controller fixed.
The undisturbed task must first satisfy the same success criterion.
Discovery, acquisition, admission, and evaluation share a prescribed task
horizon and termination rule; any terminal-quality requirement is fixed
before disturbance search.

Let $C_{k-1}$ be the currently admitted controller, which may already contain
recovery skills. At update $k$, let $\mathcal{F}_k$ be measured failures of
both $C_{k-1}$ and $\pi_0$, and let $\mathcal{Q}_{k-1}$ contain conditions enrolled in the
successful-trajectory preservation memory. The learning objective is constrained
improvement:
\begin{equation}
\begin{split}
 &\text{increase}\quad \sum_{q\in\mathcal{F}_k}Y(C_k,q),\\
 &\text{subject to}\quad Y(C_k,q)=1\quad
                   \forall q\in\mathcal{Q}_{k-1}.
\end{split}
\label{eq:contract}
\end{equation}
This is a finite-set objective, not a population guarantee. A bounded search
can fail to find an admissible improvement; then $C_k=C_{k-1}$. Retention is
measured relative to the current composite, so previously acquired recovery
skills become part of the competence protected by subsequent updates.

The bank $\bank_k=\{(r_i,g_i)\}$ stores feedback skills $r_i$ and binary
entry rules $g_i$. Runtime features $x_t\in\mathbb{R}^{d}$ may include
observations, the base action, and ordinary task-reference features; no
external disturbance schedule or label is supplied. The most recently
admitted matching entry selects one skill, which is locked for one bounded
intervention window. Outside intervention, the existing default acts:
\begin{equation}
\begin{gathered}
 \delta_t=\begin{cases}r_j(x_t,h_{j,t},v_{t-1}),&\text{active},\\
 0,&\text{otherwise},\end{cases}\\
 u_t=\mathcal{R}(\delta_t,u_{t-1}),\qquad
 a_t=\mathcal{A}(\pi_0(x_t)+u_t).
\end{gathered}
\label{eq:act}
\end{equation}
Every skill runs its feedback state forward from reset on the realized
observations and previous correction inputs, whether or not it is selected,
so entry never reinitializes that state. Here $h_{j,t}$ is skill state,
$v_{t-1}$ a previous correction input,
$\mathcal{R}$ a fixed correction limiter, and $\mathcal{A}$ the existing
action-processing interface. These maps are unchanged during learning;
zero correction preserves the base command path. The active bank starts
empty, so the initial composed controller is the bare base. Skills compete for authority rather
than having their corrections summed.

\subsection{Skill Acquisition and Entry Learning}
The acquisition stage queries the current controller to identify a failure
and searches for a recoverable response within budget $B_k$. With an ordered
disturbance parameter, increasing-strength queries and bracket refinement
provide a practical procedure for discovering failures. This does not assume a
globally monotone success boundary. Successful queries of the current controller contribute to
preservation memory; missing failures or unsuitable reference outcomes skip
the update.

Each acquired skill retains an output direction on a shared recurrent
feedback backbone:
\begin{equation}
 r_k(x_t,h_t,v_{t-1})=
 \mathcal{T}\!\left(f_\psi(x_t,h_t,v_{t-1})+D\theta_k\right).
\label{eq:skill}
\end{equation}
Here $f_\psi$ is the inherited feedback model, $D$ fixes coordinate scales,
and $\mathcal{T}$ is the interface's fixed output transformation. Search
learns $\theta_k$; entry learning determines when this skill acts. A constant
direction does not imply a constant correction: the backbone responds to
the current observation and realized history throughout intervention.
All skills receive the same interface-defined previous correction input and evolve from reset,
so their identical recurrent backbones can share one state. Acquiring a
new direction leaves earlier feedback maps and entries unchanged.

The budgeted acquisition operation is
\begin{equation}
 (r_k,\mathcal{D}_k)=
 \mathsf{Acquire}(C_{k-1},\mathcal{F}_k,B_k),
\label{eq:acquire}
\end{equation}
where $\mathcal{D}_k$ contains successful teacher trajectories with candidate
entry states. Cross-entropy search first finds a recovery at the measured
failure. A neighborhood evaluation then keeps the direction that recovers
the center condition and covers the most failures shared by the base and the
current controller. The executed feedback solution is retained directly,
without a subsequent imitation step. The selected center and at least one
further eligible training condition must recover. Teacher search may use
experimental disturbance information, which is not an input to runtime
selection. We do not require the acquisition optimizer to solve
Eq.~\ref{eq:contract} globally.

Let $\mem_{k-1}$ be the cumulative successful-trajectory memory. Its complete
recorded state histories form the exclusion set
\begin{equation}
 \negset_k=\bigcup_{\tau\in\mem_{k-1}}\{x_t:x_t\in\tau\}.
\label{eq:memory}
\end{equation}
Successful teacher onsets form the positive set $\pos_k$. Pre-entry teacher
prefixes can additionally constrain premature invocation. Distinct
successful histories at the same task remain distinct constraints; a task
identifier alone does not represent the behavior to preserve.

We use a linear entry as a minimal implementation. With standardized
features $z=S^{-1}(x-c)$, where $S$ is a diagonal matrix of per-coordinate
scales, a squared hinge loss with balanced classes fits
\begin{equation}
\begin{split}
 \min_{w,b}\;&\lambda\lVert w\rVert_2^2
 +\frac{1}{2|\negset_k|}\sum_{z\in\negset_k}[1+w^\top z+b]_+^2\\
 &+\frac{1}{2|\pos_k|}\sum_{z\in\pos_k}[1-w^\top z-b]_+^2.
\end{split}
\label{eq:entryfit}
\end{equation}
In Eq.~\ref{eq:entryfit}, $\negset_k$ and $\pos_k$ denote the standardized
versions of the two sets. After fitting, scores are recomputed as
$w^\top z$ without the fitted bias $b$, and the threshold $\beta_k$ is set
midway between the largest negative and smallest positive score:
\begin{equation}
 g_k(x)=\ind[w_k^\top S_k^{-1}(x-c_k)>\beta_k].
\label{eq:entry}
\end{equation}
An append requires a strict separating gap verified in runtime precision.
The important constraint is exclusion of successful histories, rather than
a particular classifier architecture. It addresses composition interference:
unchanged old experts can still lose competence if a new entry redirects
the states on which they previously succeeded.

\subsection{Admission Tests and Trajectory Preservation}
Form candidate $\widetilde C_k$ by appending $(r_k,g_k)$ without changing old
skills or entries. Run the candidate with autonomous selection on its
selected acquisition targets $\mathcal{A}_k\subseteq\mathcal{F}_k$ and on all
protected successful conditions. Admit only if $\mathcal{A}_k$ is nonempty,
all selected targets recover, and all protected conditions remain successful:
\begin{equation}
\begin{gathered}
 Y(\widetilde C_k,q)=1\quad
 \forall q\in\mathcal{A}_k\cup\mathcal{Q}_{k-1},\\
 C_k=\begin{cases}\widetilde C_k,&\text{checks pass},\\
 C_{k-1},&\text{otherwise}.\end{cases}
\end{gathered}
\label{eq:admit}
\end{equation}
Entry separation is checked before these rollouts. Admission requires
entry driven by observations, exact reproduction of each selected teacher,
and exact parity with reference trajectories of the current controller on all
protected conditions. Search uses an onset one tick after the simulated
disturbance; autonomous reproduction must recover that timing from observations.
Historical exclusion states and current replay
targets serve different purposes: the former constrain new selection; the
latter compare complete executable controllers from identical resets.
Accepted successful experience is retained before the next update.

\textbf{Proposition (finite-trajectory preservation).}
Consider a recorded trajectory of $C_{k-1}$ with deterministic reset and
transitions. Assume unchanged base and old skill modules, old routing, and
initial states. The augmented state includes physical state, old recurrent
histories, entry lock, task clock, previous correction inputs, and limiter
state. If the new skill cannot affect these variables before selection and
its entry is false at every recorded state, appending it reproduces the
trajectory exactly.

\emph{Proof.} Old augmented states agree at reset. If they agree at time
$t$, observations and old outputs agree. The new entry is false, so old
selection and the processed action agree. Deterministic transitions and
feedback updates preserve equality at $t+1$. Induction gives equality through
termination.\hfill$\square$

The result depends on controller composition, not on morphology, actuator
count or the optimizer that acquired a skill, which is the transferable
principle here: \emph{learn a useful feedback intervention while
constraining where it may replace existing behavior}. The tested protection
set holds previously successful trajectories of the composite, not every
baseline success, and neither entry exclusion nor replay certifies unseen
states.

\section{Experimental Protocol}
We apply the same retained-feedback acquisition, entry exclusion, and exact
admission mechanism to two control systems, on two motions each. Both active banks start empty.
The base controller and a previously prepared recurrent feedback backbone
remain fixed; only output directions and entry rules are acquired during
the reported streams. Inherited-controller preparation is excluded from
the budgets of new stages. Table~\ref{tab:backend} lists interface and budget choices.
Center search uses eight generations of eight candidates, two elites, and
a $0.5$ mean-update rate. Entry fitting uses $\lambda=5\times10^{-5}$
and a coordinatewise standardization floor of $10^{-3}$. Both interfaces
allow entry driven by observations after a four-tick warmup.

A stage is one sequential failure-discovery, acquisition, and admission
attempt. Bank after stage $k$ is the cumulative controller after the first
$k$ attempts, including earlier admitted skills. Every checkpoint is tested
on the entire fixed evaluation panel, not only its own disturbance regime.
Failed or skipped attempts leave the preceding bank unchanged.

\begin{table}[t]
\caption{Shared learning mechanism with fixed controller interfaces.
Search budgets count center and neighborhood rollouts per stage.}
\label{tab:backend}\centering\small
\setlength{\tabcolsep}{3pt}
\begin{tabular}{@{}>{\raggedright\arraybackslash}p{.25\columnwidth}>{\raggedright\arraybackslash}p{.33\columnwidth}>{\raggedright\arraybackslash}p{.34\columnwidth}@{}}
\toprule
Interface & CMU & G1\\\midrule
Base controller & Released NPMP & Released SONIC\\
Features / actions & 321 / 56 & 95 / 29\\
Learned direction & 56 dimensions & 29 dimensions\\
Search budget & $64+16\!\times\!5$ & $64+64\!\times\!7$\\
Intervention & 14 ticks ($0.42$\,s) & 20 ticks ($0.4$\,s) plus limiter tail\\
Output map & Scaled $\tanh$; action clipping & Magnitude and rate limiting\\
\bottomrule
\end{tabular}
\end{table}

\subsection{CMU Full-Reference Motion Tracking}
We use the released NPMP humanoid controller in CMU motion-tracking
simulation \cite{merel2019npmp,wagener2022mocapact}. The task follows motion
029\_07 from reference frame 241 to its natural end. It completes in
625 control ticks at $30$\,ms per tick ($18.75$\,s), without reference
padding, looping, or latent-state hold. Two successive disturbance regimes
apply horizontal root-velocity perturbations at ticks 101 and 350
($3.03$ and $10.50$\,s). Mass, friction, actuator parameters, and the base
controller remain unchanged. This first motion was exposed during
development, so its undisturbed completion establishes nominal eligibility
rather than generalization; the second stream below removes that caveat.

The same full reference is used throughout search and validation. A recovery
must complete the motion with tracking error never exceeding $0.3$ and with
a final-100-tick maximum error at most $\min(0.15,e_{\mathrm{nom}}+0.05)$,
where $e_{\mathrm{nom}}=0.03325$ is the undisturbed final-100-tick maximum
measured before disturbance search. The resulting limit, $0.08325$, is fixed
for the whole study. The tracking error combines mean absolute joint and
weighted body-position deviations from the reference. This criterion asks
for a return toward nominal tracking over the remaining task, rather than
survival to a short intermediate cutoff. The distinction matters because
endpoint survival overstates recovery: a controller can stay upright to a
truncated horizon while remaining permanently behind its reference.
Requiring the full reference and a terminal tracking bound removes that
outcome from the success count.

In Eq.~\ref{eq:skill}, the output map is $\mathcal{T}(y)=0.2\tanh(y)$ and
the coordinate scaling is $D=\operatorname{diag}(\sigma_\psi)$, with
$\sigma_\psi$ formed from the inherited log standard deviation $\ell_\psi$
as $\exp(\operatorname{clip}(\ell_\psi,-3.5,-0.2))$. The previous correction input is
the pre-action-clipping residual divided by $0.2$; the summed action is clipped to
its original range. Following center search, each of 16 directions is tested
on five training conditions: the center, and four conditions that jointly
vary onset by one tick, strength by $0.01$, and direction by $1^{\circ}$.
Multiple successful onsets and their prefixes supply entry examples. No new
imitation of the feedback network or reinforcement-learning updates are performed.

Before adaptation, 24 random offsets per regime are fixed: onset within
$\pm2$ ticks, strength within $\pm0.015$, and direction within
$\pm2^{\circ}$. After both updates are locked, the bare base and both bank
prefixes execute all 48 conditions around the discovered boundaries.
A separate rerun of the locked models checks
exact trajectory reproduction and executes the mechanism controls and every
physical development condition.

A second stream replicates this on a motion no component was tuned on.
Holding the criterion fixed and varying only the reference start frame, as
the first motion itself starts at frame 241, one undisturbed base rollout
screened each of 115 corpus clips: 10 complete their full reference, a 9\%
eligibility rate showing how much stricter this criterion is than survival
over a short window. A prespecified rule takes the first eligible clip in
sorted order from outside the twelve development cases, giving motion
026\_05 from frame 401 over 600 ticks. Timings reuse the first stream's
relative fractions, and every budget, threshold, panel construction and
control is unchanged. We call the streams CMU-A and CMU-B.

\subsection{G1 Fixed-System Simulation}
We use SONIC with the original G1 mass, friction, and PD gains and the dance
motion \texttt{dance\_in\_da\_party\_001}. All dynamics multipliers remain
one. The horizon is 496 control ticks at $20$\,ms per tick. Failure occurs for
nonfinite position, root height below $0.35$\,m, or upright-axis cosine below
$0.25$; completion reaches the horizon without those events. A horizontal
pelvis force acts for $100$\,ms with
$F=m\alpha/0.1$, where $m=35.11$\,kg and $\alpha$ is the search strength;
its impulse is $m\alpha$. The two acquisition regimes are near phase
$0.4$ and $0.75$. In the completed reference run their selected strengths
are $2.25$ and $2.6$, corresponding to impulses of approximately $79.0$
and $91.3$\,N\,s. These are simulation conditions, not a prescription for
manual physical pushes.

The inherited G1 feedback network was prepared by distillation and
22 warm-policy PPO updates. That preparation includes this first motion and
development experience near phase $0.75$; it is not charged as new-stage
learning. Each new G1 skill adds a searched 29-dimensional output offset
with scale $0.08$ to this frozen feedback network: $D=0.08I$ and
$\mathcal{T}$ is the identity in Eq.~\ref{eq:skill}. No new feedback-network
gradient updates occur during these stages.

After the center search, 64 directions are evaluated on seven conditions:
center, $\pm1^{\circ}$ direction, $\pm2$ control ticks in onset, and
$\pm0.037$ in strength. Corrections last 20 ticks with a magnitude
bound of $0.3$ and a rate limit of $0.05$ per tick. All feedback models receive
the same previous limiter-executed correction normalized by $0.5$.
Here
$\mathcal{R}$ bounds and rate-limits the correction, including its tail,
before summation with SONIC; $\mathcal{A}$ leaves that summed command
unchanged before the existing low-level stack.

Before adaptation, 48 random perturbation offsets are fixed, 24 per regime:
onset within $\pm2$ ticks, strength within $\pm0.06$, and direction within
$\pm2^{\circ}$. They are applied around the discovered centers and evaluated
only after all adaptation is locked. The audit checks physical-condition
overlap with fitting and discovery data. Each of SONIC, both bank prefixes,
latest-skill substitution, and zero-offset warm feedback executes all
48 conditions. Two further complete search-seed repetitions use the same algorithm and
budgets, each starting from an empty bank and evaluating offsets around its
own discovered boundary. That stream and its two repetitions are reported
separately; their conditions are not pooled.

A second G1 stream uses the motion \texttt{forward\_lunge\_R\_001} over a
398-tick horizon. SONIC completes 12 of the 13 available reference motions
undisturbed; among those, a prespecified rule excludes the admitted motion,
requires a horizon between 350 and 600 ticks so both phase regimes leave
adequate remaining task, and takes the first in sorted order. This stream is
also a transfer test, because the inherited feedback backbone was prepared
on the first motion alone. Panel construction, budgets and controls are
unchanged. We call the four streams CMU-A, CMU-B, G1-A and G1-B, the B
streams being the ones nothing was tuned on.

\subsection{Metrics, Mechanism Controls, and Audits}
The following definitions apply to all four streams.
\emph{Latest-skill substitution} uses the same learned entries and expert
weights, but executes the newest skill whenever any entry fires. It tests
retention under replacement, not a separately optimized shared-policy
continual learner. \emph{Zero-offset warm feedback} retains the learned
entry and inherited feedback backbone while removing the searched offsets.
The bare base supplies the zero-intervention control. \emph{No exclusion}
and \emph{minimal exclusion} refit only the final bank's entry rules. The
first keeps teacher prefixes recorded before entry as the sole negatives; the second
adds back, one admission replay at a time, those protected trajectories the
replay shows to have changed, stopping when parity holds. Both preserve the
searched directions, teacher positives, objective and solver of the admitted
run, and neither is re-admitted: each is reported as a controller the
admission test would have judged, not as a method variant. Discovery and
search are not re-run for either.

On a panel we report paired counts against a reference controller $C$:
gains are the conditions $C$ fails and the final bank recovers, losses the
reverse. A strict stage acquisition must succeed after the stage while failing under
both the bare reference and the preceding bank, with post-disturbance entry.
Restoring an earlier regression is reported separately from acquiring a
new recovery. Forgetting counts an earlier bank success lost at a later
checkpoint. We retain failed and skipped stages in the record.
The reference run failed our original screen, which allowed no harmed
condition at any stage. We relaxed that screen post hoc to at most two
distinct harmed conditions and then fixed the relaxed tolerance in advance
for both repetitions.

We report $G$ and $L$ rather than significance levels. Whenever $L=0$, an
exact two-sided paired sign test is a deterministic function of the gain
count, $p=2^{1-G}$, and so carries no information the counts do not already
carry. More importantly, the 48 conditions of a panel are local
perturbations of one learned system, not independent learning trials, so a
small $p$ would describe the sampled neighborhood rather than the
reliability of the learning procedure. The two complete repetitions, not the
panel size, are our evidence about reliability.

A CMU stream costs about 540 rollouts and a G1 stream 1,475 logical
trajectories, of which two thirds reuse a cached deterministic prefix; the
refit arms add no search and reuse the recorded conditions. These figures
exclude inherited controller preparation and are not a matched-budget
comparison. Artifact audits check source and trace hashes, fixed dynamics, teacher
provenance, entry timing, memory and data separation; an independent replay
checks recorded observations, feedback states, expert outputs and executed
corrections; and exact arithmetic checks use the executed inference backend
rather than treating CPU and GPU roundoff as interchangeable.

\section{Results}
\begin{figure*}[t]
\centering
\includegraphics[width=0.88\textwidth]{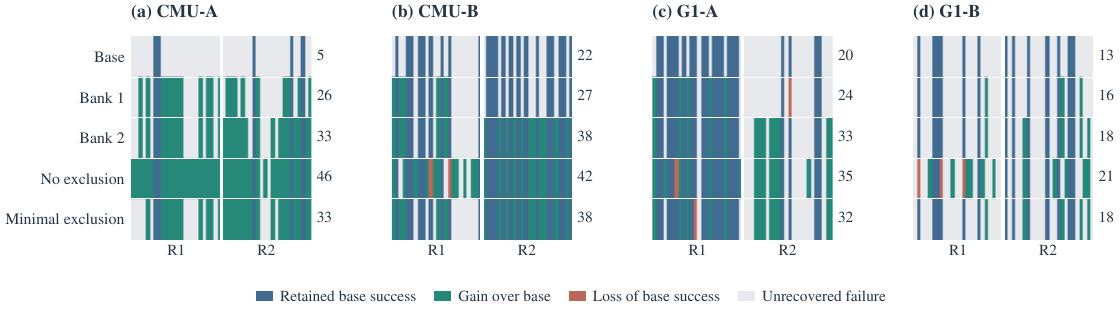}
\caption{Per-condition outcomes on each stream's 48 withheld perturbations, grouped
by acquisition regime (R1, R2), with totals at the right. The last two rows
keep the final bank's skills and refit only its entry rules.}
\label{fig:outcomes}
\end{figure*}

\begin{figure}[t]
\centering
\includegraphics[width=\columnwidth]{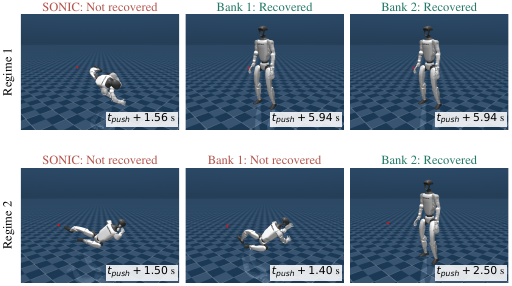}
\caption{Two sequential G1-A acquisitions, one per regime, under the base controller
and the two banks on the same disturbance. Poses are the last recorded
pre-action states, labeled by the audited episode outcome; times are
relative to the push.}
\label{fig:qualitative}
\end{figure}

\subsection{Sequential Recovery Acquisition}
\begin{table}[t]
\caption{Every comparator against the final bank of its own stream, over 48 withheld
perturbations per panel. Each cell reads \emph{successes
(bank-only/comparator-only)}: conditions only that stream's final bank
recovers, then conditions only the comparator recovers. CMU-B and G1-B are the motions
nothing was tuned on; the exclusion arms refit only entry rules.}
\label{tab:main}\centering\small
\setlength{\tabcolsep}{3.4pt}
\begin{tabular}{@{}lrrrr@{}}
\toprule
Comparator & CMU-A & CMU-B & G1-A & G1-B\\\midrule
Base & 5 (28/0) & 22 (16/0) & 20 (13/0) & 13 (5/0)\\
Latest skill & 36 (0/3) & 36 (3/1) & 30 (3/0) & 15 (3/0)\\
Warm, zero offset & 3 (30/0) & 33 (6/1) & 28 (7/2) & 14 (5/1)\\
No exclusion & 46 (0/13) & 42 (4/8) & 35 (1/3) & 21 (6/9)\\
Minimal exclusion & 33 (1/1) & 38 (0/0) & 32 (1/0) & 18 (0/0)\\\midrule
\textbf{Final bank} & \textbf{33} & \textbf{38} & \textbf{33} & \textbf{18}\\
Protected conditions & 22 & 21 & 23 & 17\\
\quad kept without exclusion & 22 & 19 & 23 & 11\\
\quad reproduced exactly & 8 & 10 & 13 & 6\\
\bottomrule
\end{tabular}
\end{table}
Both CMU stages are admitted. The first reproduces four selected training
recoveries while exactly replaying seven protected trajectories; the second
reproduces two further recoveries while replaying 20. Table~\ref{tab:main}
reports the withheld panel. Two features of it are not visible in the
totals (Fig.~\ref{fig:outcomes}). The opening update already transfers
across regimes, nine of its 21 gains falling at a disturbance timing it
never searched, and the closing update then leaves the first regime
untouched and places every gain in the second. That is the division of
labor the entry constraint is meant to produce.
Figure~\ref{fig:qualitative} shows the same progression on G1-A.
Independent simulation reproduces all 144
main-panel trajectories exactly. On all 27 distinct development conditions,
success increases from 14 to 21 and then 23, again without base-success
losses or forgetting.

Both CMU-B stages are admitted as well, and its panel repeats the pattern
from a very different starting point: the base already recovers 22 of 48
conditions there, yet two updates still add 16 more without surrendering
one. Acquisition therefore does not depend on the base controller being
near-helpless, and it survives a motion that no part of the procedure was
tuned on.

Both G1 stages are admitted in the reference run. The first teacher covers
all five eligible training failures in its seven-condition neighborhood; the
second covers six of seven. Admission reproduces the selected teachers and
preserves the current successful references. On the withheld panel the
second bank adds nine successes over the first: eight new recoveries and the
restoration of the single condition the first bank had harmed. Strict stage
acquisitions, which exclude conditions already recovered by the bare base or
the preceding bank, are five and eight. All 23 final memory tasks succeed
and nominal behavior is preserved.

G1-B admits both stages as well, so the backbone does transfer off the
motion it was prepared on, but weakly: five gains against thirteen on G1-A,
from a base that starts lower. Every protected condition survives. We read
this as evidence that the acquisition mechanism is not motion-specific while
the inherited backbone partly is.

The two prescribed G1 repetitions yield $25/48\rightarrow29/48$ and
$25/48\rightarrow30/48$ with no losses, each retaining all 16 memory
conditions, but each admits only its first stage. Both stop at the same
place and for the same structural reason: a stage may target only conditions
the base and the current bank both fail, so a condition the bank itself
harmed can never become a target, and the stage is skipped before teacher
search. Sec.~\ref{sec:discussion} develops this. We regard the stall, rather than
the reference run's totals, as the result that locates the remaining work.

\subsection{Skill Composition and Retention}
Table~\ref{tab:main} collects the comparators. Deleting the searched
directions while holding the learned entries fixed drops CMU-A recovery
below the bare base, so firing the inherited backbone with no searched
direction is worse than not intervening at all and an entry rule is worth no
more than the skill it admits; on the other three streams the same deletion
costs far less, because there the backbone and entry rule already recover without any
searched offset. Substituting the newest skill at every entry beats
our selector only on CMU-A, where the conditions it adds are ones the bank
routes to the earlier skill although the newer direction recovers from the
same state; on the other three streams the bank is ahead by a similar
margin. The comparison is close everywhere, so we claim no routing
advantage in either direction. Component effects are system dependent and should not
be read off the aggregate gain.

\subsection{The Cost of Exact Preservation}
The constraint that distinguishes this method is the exclusion of recorded
successful states from a new entry region. Two arms in Table~\ref{tab:main}
isolate what it buys and what it costs; both reuse the final bank's searched
directions and teachers and refit only the entry rules.

Dropping the successful-trajectory negatives, leaving the teacher prefixes
as the only negatives, raises recovery on all four streams. No arm ever
triggers before its disturbance, so the prefixes alone preserve entry
timing. What the memory negatives do is suppress entry: the admitted entry
never fires on about a quarter of each panel, while the unconstrained entry
fires throughout. That suppression is indiscriminate, because the constraint
is fitted on states resembling past successes, which is a different question
from whether intervening here would succeed now. It therefore charges where
the base controller was failing and protects where it was succeeding, and
that balance is not a constant of the method. Summed over the four streams
it forgoes 33 recoveries and saves 11 withheld conditions, but the split is
uneven: CMU-A is pure cost, 13 recoveries for no harm avoided, while G1-B
forgoes nine and saves six.

What it delivers reliably is exact behavior, and the B streams expose the
failure it exists to prevent. Without the constraint eight protected
conditions across those streams stop completing at all, rather than merely
taking a different route, six of them on G1-B where more than a third of the
protected set collapses. Of 83 protected trajectories, 46 survive unchanged
only because of the constraint. A protocol scoring retention by success
alone would call it costless on CMU-A and useless on G1-A, and would be
wrong both times.

The cost is not an artifact of excluding too much. The arm that excludes only the necessary trajectories
fits the entry on teacher prefixes, replays the protected conditions, and
adds back only the trajectories the replay shows to have changed, repeating
until parity holds. It converges in two rounds per stage on all four streams
and retains 53--84\% of the states the method excludes, yet returns none of
the gap on any of them; on both B streams it reproduces the admitted
controller condition for condition. The explanation is structural. A protected
trajectory that can be dropped is one the entry never fired on, so it was
never constraining; the trajectories that must be kept are exactly those
where entry wanted to act. Within this family of entry rules, exact
preservation has no cheaper purchase.

These arms also bound what better routing could return. The acquired skills
reach conditions that the controllers admitted from them decline, 13 on
CMU-A, so the gap belongs to entry rather than to what was learned and its
size is measured rather than asserted. Closing it needs a selector that
decides differently rather than more often: a selector separating the eight
conditions worth entering on CMU-B from the four worth declining would take
that headroom without weakening the guarantee, whereas entering everywhere,
as these arms do, cannot.

\section{Discussion and Limitations}
\label{sec:discussion}
One acquisition mechanism carries four streams over two systems, though the
inherited backbone is partly motion specific: G1-B admits both stages but
gains far less than G1-A, the motion that backbone was prepared on. Nothing
here demonstrates transfer of a learned skill between plants.

Exactness and coverage turn out to be coupled through the same states. A
protected trajectory is droppable only when the entry never fired on it, so
dropping it frees nothing; the trajectories that bind are those the entry
wanted to act on. Tuning the exclusion policy is therefore not the remedy,
and the component to change is the selector: retention is enforced per skill
while recovery is a property of the bank, so a selector ranking admitted
skills by expected recovery would still satisfy the proposition and has a
measured 13 conditions of headroom on CMU-A. Preservation is also exact only
on the protected trajectories, and entry outside that memory can still
generalize poorly: G1-A's intermediate loss follows a late erroneous entry,
repaired by its second bank. On 12 previously exposed G1 challenge
conditions the final bank has two gains and one loss against SONIC, while
all 25 further protection conditions hold.

Both G1 repetitions stall for a related reason that belongs to our
eligibility rule rather than to the search. A stage may target only
conditions the base and the current bank both fail, which is what makes its
gain attributable to it; a condition the bank harms is by construction one
the base still handles, so it can never become a target. The rule that keeps
credit clean thus leaves the stream blind to the regressions it causes, and
both repetitions skip their second stage on exactly that ground. Reliable
repeated learning needs a second kind of stage aimed at an inappropriately
firing entry, which is a different problem from acquiring a missing skill.

The evidence covers short streams, simulator resets, inherited feedback
backbones, and one intervention window per episode. CMU evaluates one
nominally valid complete reference at two disturbance timings; neither
system establishes broad robustness across unseen motions or physics distributions.
Each exclusion set contains the previous one, so admissible entry rules can
only shrink as a stream lengthens and the acquisition rate should saturate;
our streams are too short to locate that point, and long-stream memory
compression is untested. Finite task success is also not an asymptotic
stability guarantee, and on G1-A's jointly successful conditions mean tracking
RMSE rises by $0.0016$\,rad, so retained completion need not imply unchanged
motion quality. Real-robot learning remains unvalidated.

\section{Conclusion}
We treated sequential recovery as an admission problem rather than an
optimization problem: a skill enters a learned controller only after the
composed policy reproduces the new recovery from observations and replays
every protected trajectory exactly. Constraining entry by accumulated
successful histories, instead of freezing parameters alone, is what makes
that test meaningful when a selector can redirect a frozen skill, and a
finite-trajectory condition states when it preserves the previous controller
exactly. One mechanism adds recoveries on four streams over two systems,
including two motions nothing was tuned on, without surrendering a base
success anywhere. We then measured what the guarantee costs, which to our
knowledge has not been reported: 33 recoveries forgone, against 11 withheld
and eight protected conditions saved and 46 of 83 trajectories held
identical. A control that excludes
only what the admission replay proves necessary shows the cost is intrinsic,
because the trajectories that must be excluded are the ones that were
suppressing entry. The same arms bound the remaining headroom, since
the acquired skills reach conditions the admitted controllers decline: what
limits those controllers is where they may act rather than what they
learned. Two concrete components follow. One is a stage directed at repair, since a
stream that may target only failures shared with the base cannot repair a
regression it caused itself. The other is a selector ranked over the
admitted bank, since deciding more often returns nothing while deciding
differently has measured room. Physical validation remains the eventual
test, but it is not the binding constraint here; the reliability of repeated
acquisition is.

\section*{Acknowledgment}
OpenAI Codex assisted software development, experiment orchestration,
artifact checking, drafting and figures. Scientific claims are tied to the
recorded experiment and audit artifacts.
{\scriptsize
\bibliographystyle{IEEEtran}
\bibliography{references}}
\end{document}